\documentclass{bmvc2k}

\usepackage{booktabs}
\usepackage{xspace}
\usepackage{latexsym}
\usepackage{amssymb}
\usepackage{amsmath}
\usepackage{amsthm}
\usepackage{booktabs}
\usepackage{enumitem}
\usepackage{graphicx}
\usepackage{color}

\usepackage[table]{xcolor}
\usepackage{multirow}
\usepackage{graphicx}
\usepackage{subcaption}
\usepackage{xcolor}
\usepackage{placeins}
\title{Not All Negatives Are Equal: Query-Adaptive Routing for Few-Shot Vision-Language Models}

\addauthor{Sriram Mandalika}{sriram.mandalika@student.hpi.uni-potsdam.de}{1}

\addinstitution{
 Hasso Plattner Institute,\\
 University of Potsdam\\
 Potsdam, Germany
}

\runninghead{Mandalika}{Not All Negatives Are Equal}

\begin{document}

\maketitle

\begin{abstract}
Few-shot adaptation of vision-language models remains fundamentally limited by how negative class signals are handled at inference. Existing methods apply uniform negative suppression across all queries, ignoring that the most damaging confusions are query-specific and shift with support-set geometry. We introduce SCAN (Selective Confusion-Aware Negatives), a framework that addresses this gap through three targeted contributions. At inference, query-adaptive negative routing restricts suppression to the top-K most confusable classes per query, requiring zero additional parameters. Generic negative text templates are replaced with LLM-bootstrapped contrastive prompts that describe discriminative attributes between confusable class pairs, sharpening the textual decision boundary where it matters most. A parameter-free adaptive fusion weight estimated from support-set Fisher discriminability removes the need for manual tuning of the vision-language trade-off. Evaluated across 11 standard benchmarks, SCAN consistently outperforms prior prompt-based and adapter-based methods by an average of 4.61\% at 16-shot, with gains of up to 7.70\% on fine-grained datasets where inter-class confusion is most severe. SCAN also generalizes strongly under distribution shift, improving by 2.95\% on average across four ImageNet OOD variants, and maintains robust performance under significant label noise, with accuracy under 50\% label corruption still exceeding the clean baseline of the strongest competing method.
\end{abstract}

\section{Introduction}
\label{sec:intro}

In many real-world settings, such as medical diagnosis, remote sensing, or wildlife monitoring, deep models are expected to learn from just a handful of labelled examples. While deep learning has shown remarkable success in visual recognition \cite{Redmon2015YouOL, Russakovsky2014ImageNetLS}, these gains largely rely on the availability of large-scale labelled datasets \cite{Deng2009ImageNetAL, Lin2014MicrosoftCC}, which are difficult or impractical to obtain in specialized domains. This makes few-shot learning a critical and unsolved challenge for deploying visual models in-the-wild.

Vision-language models (VLMs) such as CLIP~\cite{Radford2021LearningTV} and ALIGN~\cite{Jia2021ScalingUV}, pretrained on web-scale image-text pairs, offer strong zero-shot capabilities \cite{Li2021SupervisionEE, Radford2021LearningTV, Yu2022CoCaCC, Zhai2021LiTZT}. These models provide a powerful foundation for few-shot transfer. However, the current few-shot adaptation strategies fall into two broad categories: adapter-based fine-tuning \cite{Gao2021CLIPAdapterBV, Li2023GraphAdapterTV, Yu2022TaskRF, Zhang2022TipAdapterTA} and prompt-based tuning \cite{Zhang2024ConceptGuidedPL, Zhou2022ConditionalPL, Zhou2021LearningTP, Tian2023ArGueAP, Shu2022TestTimePT}. Both approaches struggle under low supervision, especially with fine-grained classes, due to their over-reliance on fixed prompts or shallow adaptation modules that cannot adequately encode task-specific variation.

To address this, we propose \textbf{SCAN}, \textbf{S}elective \textbf{C}onfusion-\textbf{A}ware \textbf{N}egatives, a few-shot vision-language adaptation framework built around the insight that negative signals should be query-specific rather than uniform. Prior negative learning methods~\cite{Zhang2024EnhancingVL}\cite{SimNL} apply the same negative penalty to every test image, regardless of which classes that particular image is actually likely to be confused with. SCAN instead identifies the most confusing classes for each query at inference time and routes negative suppression selectively toward them, concentrating the penalty where it matters most. This query-conditioned routing sharpens per-query decision boundaries without introducing any additional parameters or training overhead, and is compatible with any adapter-based few-shot pipeline.

Beyond the routing mechanism, SCAN addresses two further weaknesses in how current methods handle negative supervision. For textual negatives, we replace generic class-level templates with LLM-generated pairwise contrastive descriptions, where each class is characterized against its most visually similar neighbors rather than described in isolation, providing grounded discriminative language supervision that goes well beyond simple negations of class names~\cite{Pratt2022WhatIA}. For modality fusion, we introduce a parameter-free adaptive weighting mechanism that estimates the optimal balance between visual and textual branches from the geometry of the support set alone, removing the need to manually tune the fusion hyperparameter and allowing the model to naturally adapt to datasets where one modality is more discriminative than the other.

We evaluate SCAN on 11 standard benchmarks spanning general recognition, fine-grained categorization, and cross-dataset transfer. Our experiments show that SCAN consistently outperforms prior few-shot adaptation methods, with the largest gains on fine-grained datasets where inter-class confusion is highest and selective negative routing contributes most. Each contribution is validated through a controlled ablation, and the full model achieves the best results across all evaluated settings.

Our key contributions are as follows:
\vspace{-0.3em}
\begin{itemize}
    \item We propose SCAN, a few-shot vision-language adaptation method built around query-conditioned negative routing, where negative suppression is concentrated on the classes each test image is most likely to be confused with rather than applied uniformly across all classes.

    \item We introduce LLM-bootstrapped pairwise contrastive prompts as negative textual supervision, replacing generic class templates with targeted descriptions that characterize each class against its most visually similar neighbors, generated entirely offline before training.

    \item We design a parameter-free adaptive modality fusion mechanism that infers the optimal weighting between visual and textual branches from support-set geometry alone, naturally favoring vision when visual features are discriminative and text when class names are semantically close.
\end{itemize}

\section{Related Works}
\vspace{-0.7em}
\noindent\textbf{Efficient Adaptation for VLMs.} Prompt-based methods adapt frozen text encoders by learning task-specific prompts. CoOp~\cite{Zhou2021LearningTP} learns fixed prompts via gradient descent, while CoCoOp~\cite{Zhou2022ConditionalPL} generates instance-aware prompts conditioned on image features. ProGrad~\cite{Zhu2023PromptAlignedGF} mitigates overfitting through gradient projection, and TPT~\cite{Shu2022TestTimePT} enhances robustness via inference-time prompt refinement. A complementary line of work enriches class-level textual supervision using large language models. CuPL~\cite{Pratt2022WhatIA} prompts GPT-3 to generate category-specific visual descriptions that capture attributes well beyond simple class names, and VisDesc~\cite{Menon2022VisualDA} similarly uses LLM-generated visual features as zero-shot classifiers. While these works demonstrate that richer language supervision improves recognition, they focus exclusively on positive class descriptions and do not model discriminative relations between visually similar classes. SCAN builds on this direction by generating pairwise contrastive descriptions that explicitly characterize each class against its nearest semantic neighbors. In parallel, adapter-style methods retain frozen backbones and inject lightweight modules into the vision stream. CLIP-Adapter~\cite{Gao2021CLIPAdapterBV} uses residual adapters for supervised tuning, while Tip-Adapter-F~\cite{Zhang2022TipAdapterTA} constructs a non-parametric cache for training-free alignment. Graph-based extensions like Dual-Knowledge Graph tuning~\cite{Anonymous2023DKG} exploit semantic relations between classes. TaskRes~\cite{Yu2022TaskRF} introduces task-specific residual routing, and Sus-X~\cite{Udandarao2023SuSX} achieves plug-and-play adaptation with zero training cost. Most of these methods require manual tuning of a scalar weight that balances visual and textual branch contributions, whereas SCAN estimates this weight directly from support-set geometry without any additional parameters.

\noindent\textbf{Few-Shot Learning.} Few-shot learning (FSL) aims to recognize unseen classes from limited labeled examples. Meta-learning approaches such as Matching Networks~\cite{Vinyals2016MatchingNF}, Prototypical Networks~\cite{Snell2017PrototypicalNF}, and MAML~\cite{Finn2017ModelAgnosticMF} simulate $N$-way $K$-shot tasks to learn transferable inductive biases. Meta-Baseline~\cite{Chen2021MetaBaseline} and Jiang et al.~\cite{Jiang2024BoostingMetaTraining} enhance prototype quality via base-class knowledge. DeepEMD~\cite{Zhang2020DeepEMD} uses optimal transport for metric learning, while transductive models like UNEM~\cite{Smith2023UNEM} refine class prototypes with unlabeled queries. Other innovations include attentive regularization~\cite{Li2024AttentiveFeatureRegularization}, graph-based propagation~\cite{Liu2021TransductivePropagationNetwork}, and domain-adaptive methods~\cite{Gupta2024FMVP, Kumar2025CrossViewNearestNeighbor, Patel2024FeatWalk}. In vision-language FSL, CoOp~\cite{Zhou2021LearningTP} and CoCoOp~\cite{Zhou2022ConditionalPL} learn task-specific prompts, Tip-Adapter-F~\cite{Zhang2022TipAdapterTA} enables training-free adaptation via feature caching, and SimNL~\cite{SimNL} extends this with dual-branch modeling of positive and negative class prototypes.

\noindent\textbf{Negative Learning.} Negative learning aims to improve class separation by explicitly modeling "what a class is not." Early works use complementary labels to train on classes an instance does not belong to~\cite{Ishida2017LearningFC}, and Yu et al.~\cite{Yu2017LearningWB} show that negative prototypes reduce spurious correlations in prototype-based classifiers. DualCoOp~\cite{Sun2022DualCoOpFA} extends this idea to prompt tuning by learning both positive and negative prompts to suppress confusing classes. SimNL~\cite{Zhang2024EnhancingVF} builds on this by mining task-specific negatives from CLIP similarity and aligning dual visual-textual branches. Recent advances explore more expressive negative generation: DeltaAug~\cite{DeltaAug2024} synthesizes hard negatives via embedding arithmetic, DualAdapter~\cite{DualAdapter2024} injects positive-negative adapter paths, and SCHaNe~\cite{SCHaNe2024} applies supervised contrastive learning with hard negatives. NAS~\cite{Wang2024NAS} introduces token-level visual negatives during pretraining, NtUA~\cite{BMVC2024NtUA} uses hard negative mining for noise robustness in few-shot ImageNet, and MDPI~\cite{MDPI2024HardNegatives} integrates hard negatives into cross-modal contrastive training. Despite this progress, all existing approaches construct a fixed set of negative classifiers that are applied identically to every test query, without accounting for the fact that different images are confused by fundamentally different classes. SimNL~\cite{Zhang2024EnhancingVF}, APE~\cite{Zhou2022LargeLM}, and NegCLIP~\cite{Yuksekgonul2022WhenAW} all share this query-agnostic design, which dilutes the negative signal across classes that are irrelevant for a given test image. SCAN addresses this gap through selective negative routing that concentrates the penalty toward the classes each query is most likely to be confused with, rather than distributing it uniformly across the full label space.

\section{Methodology}
\subsection{Problem Formulation}

We consider the standard $N$-way $K$-shot classification task in a vision-language setting. Each episode comprises a support set $\mathcal{S}=\{(x_i,y_i)\}_{i=1}^{N\times K}$ and a query set $\mathcal{Q}=\{(x_j,y_j)\}_{j=1}^{N\times Q}$, sampled from $N$ novel classes. Images $x$ are mapped to $d$-dimensional features $v=f_v(x)$ by CLIP's frozen visual encoder, while class names $c$ yield text embeddings $t=f_t(c)$ via its frozen text encoder. Our goal is to learn lightweight adaptation modules, without updating $f_v$ or $f_t$, that refine these base embeddings into discriminative class prototypes and hard negatives. Queries are then classified by cosine similarity to the adapted prototypes.

\begin{figure*}[t]
    \centering
    \includegraphics[width=0.9\linewidth]{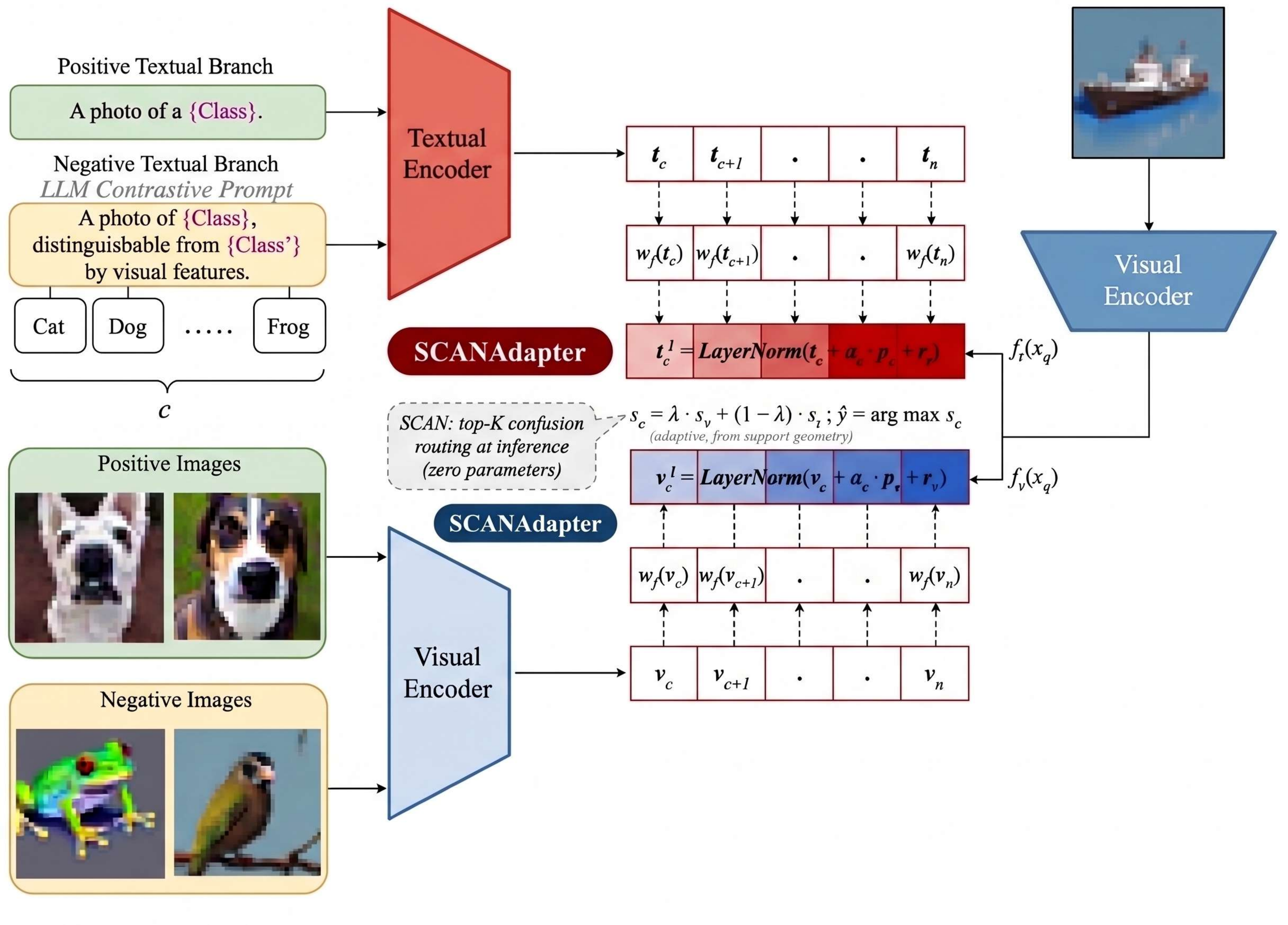}
    \caption{\textbf{Illustration of the SCAN architecture,} integrating dual-branch vision-language adaptation using learnable prompt prediction, residual-enhanced prototypes, and cross-modal fusion. Both textual and visual features are aligned and contrasted with query features through cosine similarity.}
    \label{fig:scan_arch}
\end{figure*}

\subsection{Core Learning Objectives}
\label{sec:core_objectives}

We incorporate dual positive and negative objectives. The positive branch aligns queries with their correct class prototypes via vision-language cache similarities, while the negative branch penalizes proximity to semantically similar but incorrect classes.

\subsubsection{Positive Prototype Alignment}

Given query $q \in \mathbb{R}^d$, the fused text weight $\mathbf{w}_c^+ = 0.45\,\tilde{t}_c^{\mathrm{tmpl}} + 0.55\,\tilde{t}_c^{\mathrm{cupl}}$ combines adapted template and CuPL embeddings. The positive score for class $c$ is:
\[
s_c^+ = 100\cdot(q^\top \mathbf{w}_c^+) + \alpha \sum_{i \in \mathcal{S}} \exp\!\left(-\beta\bigl(1 - q^\top \tilde{k}_i\bigr)\right)\tilde{V}_{ic},
\]
where $\alpha$ and $\beta$ are the cache scale and kernel sharpness, $\tilde{k}_i$ are the adapted visual cache keys, and $\tilde{V}_{ic}$ are instance-reweighted soft labels that upweight support samples closer to their class prototype. Training minimizes:
\[
\mathcal{L}_{\mathrm{pos}} = -\log \frac{\exp(s_y^+)}{\sum_{j=1}^N \exp(s_j^+)}.
\]

\subsubsection{Negative Learning Branch}

Let $\mathbf{w}_c^-$ be the adapted negative text weight and $k_i^-$ the pseudo-negative visual cache keys, constructed by averaging features of classes dissimilar to $c$. The negative score is:
\[
s_c^- = 15\cdot(1 - q^\top \mathbf{w}_c^-) + \alpha \sum_{i \in \mathcal{S}} \exp\!\left(-\beta\cdot q^\top k_i^-\right) V_{ic}.
\]
A hard-negative hinge loss further sharpens boundaries by penalizing queries within margin $m$ of the $k$ globally hardest negative prototypes $\mathcal{N}$:
\[
\mathcal{L}_{\mathrm{hn}} = \frac{1}{|\mathcal{N}|} \sum_{n \in \mathcal{N}} \max\!\left(0,\, m - \cos(q, z_n^-)\right).
\]

\subsubsection{Contrast with Instance-Level Contrastive Learning}

Unlike SimCLR~\cite{Chen2020Simple} and MoCo~\cite{He2020Momentum}, which rely on large negative sample banks, our method performs class-level alignment using task-specific hard negatives from the support set, making it far more data-efficient under the few-shot constraint.

\subsection{Modality-Specific Branch Designs}
\label{sec:branch_architecture}

\subsubsection{Textual Branch}

\paragraph{Prompt Prediction and Style Conditioning.} We adapt frozen CLIP text embeddings via a learnable style bank of $S$ vectors $\{s_i \in \mathbb{R}^d\}_{i=1}^S$ and a two-layer MLP $f_\phi: \mathbb{R}^d \to \mathbb{R}^S$. Given the frozen embedding $t_c$, attention weights $a_{c,i} = \exp(\tilde{a}_{c,i}) / \sum_j \exp(\tilde{a}_{c,j})$ are computed where $\tilde{a}_c = f_\phi(t_c)$, yielding a prompt token $p_c = \sum_i a_{c,i} \cdot s_i$. The enhanced prototype is $t_c' = t_c + p_c$.

\paragraph{Cross-modal Coordination.} The prompt-enhanced embedding $t_c'$ attends to the support visual features $V \in \mathbb{R}^{|\mathcal{S}|\times d}$ via cross-attention:
\[
\tilde{t}_c = \mathrm{CrossAttn}(q = t_c',\, k = V,\, v = V),
\]
grounding each text prototype in task-relevant visual evidence from the support set.

\paragraph{LLM Contrastive Prompts for Negative Supervision.} Generic negative templates such as "not a photograph of a \{CLASS\}" fail to capture which visual attributes distinguish a class from its nearest neighbors. For each class $c$, we identify its top-$k$ most visually confusable classes in CLIP embedding space and prompt an LLM to generate descriptions of the form "a photo of $c$, distinguishable from $c'$ by [visual attribute]" for each confuser $c'$. Generated offline with no additional parameters and stored in the same format as CuPL prompts, these embeddings form $\mathbf{w}_c^-$ and provide richer discriminative signal than template-based negatives.

\subsubsection{Visual Branch}

For each class $c$, the prototype $v_c$ is the mean of the CLIP-encoded support features $\{x_i \mid y_i = c\}$, enhanced with a learnable residual token $r_c \in \mathbb{R}^d$:
\[
v_c' = v_c + r_c, \quad \hat{v}_c = \mathrm{LayerNorm}(v_c'), \quad \tilde{v}_c = W_v\,\hat{v}_c,
\]
where $W_v \in \mathbb{R}^{d \times d}$ is initialized as identity. The per-shot expansion of $\tilde{v}_c$ yields the adapted cache keys $\tilde{k}_i$ used in positive branch scoring.

\subsection{Selective Confusion-Aware Negatives}
\label{sec:scan}

Prior negative learning methods~\cite{Zhang2024EnhancingVF, Zhu2023APE, Yuksekgonul2022WhenAW} apply a fixed negative classifier to every test query, distributing the penalty uniformly across all classes regardless of which ones a particular image is actually confused about. We introduce \textbf{SCAN} (Selective Confusion-Aware Negatives), which routes negative suppression toward the classes each query is most likely to be confused with. The confusion set $\mathcal{N}(q)$ contains the $k$ classes with the highest positive score:
\[
\mathcal{N}(q) = \mathrm{TopK}_k\left\{c \mid s_c^+\right\}.
\]
A per-query confusion weight, restricted to $\mathcal{N}(q)$, is computed as:
\[
\beta_c(q) = \begin{cases} \dfrac{\exp(s_c^+ / \tau)}{\displaystyle\sum_{c' \in \mathcal{N}(q)} \exp(s_{c'}^+ / \tau)} & c \in \mathcal{N}(q) \\[6pt] 0 & \text{otherwise,} \end{cases}
\]
where $\tau$ controls the sharpness of the confusion profile. The routed negative scores and final prediction are:
\[
\hat{s}_c^- = s_c^- \cdot \beta_c(q), \qquad s_c = \lambda\, s_c^+ + (1 - \lambda)\,\hat{s}_c^-, \qquad \hat{y} = \arg\max_c\, s_c.
\]
SCAN adds no parameters and is applied only at inference, making it compatible with any training procedure.

\subsection{Training and Optimization}
\label{sec:training_optimization}

\paragraph{Unified Loss Formulation.} We optimize a composite loss:
\[
\mathcal{L} = \mathcal{L}_{\mathrm{pos}} + \mathcal{L}_{\mathrm{rank}} + 8\,\mathcal{L}_{\mathrm{consist}} + \mathcal{L}_{\mathrm{hn}} + \gamma\,\|\theta_{\mathrm{attn}}\|_2^2,
\]
where $\mathcal{L}_{\mathrm{rank}}$ enforces margin constraints between queries and class prototypes, $\mathcal{L}_{\mathrm{consist}}$ penalizes drift of the adapted text weights from the original CLIP embeddings, and $\gamma\,\|\theta_{\mathrm{attn}}\|_2^2$ regularizes the cross-modal attention parameters. Optimization uses AdamW with cosine annealing; negative branch parameters are trained at five times the base learning rate.

\paragraph{Adaptive Modality Fusion.} Rather than tuning $\lambda$ by grid search, we estimate it from support-set geometry with no additional parameters. Let $\sigma_v$ be the visual Fisher discriminability (between-class to within-class variance ratio) and $\sigma_t = 1 - \bar{\rho}_t$ the textual discriminability, where $\bar{\rho}_t$ is the average pairwise cosine similarity among class text weights. The fusion weight is:
\[
\lambda = \mathrm{clip}\!\left(\frac{\sigma_v}{\sigma_v + \sigma_t},\, \lambda_{\min},\, \lambda_{\max}\right).
\]
This naturally increases $\lambda$ when visual features are discriminative and reduces it when class names are semantically close. The same $\lambda$ is used throughout training and inference.

\paragraph{Inference Strategy.} Positive and negative scores $s_c^+$ and $s_c^-$ are computed as in Section~\ref{sec:core_objectives}. SCAN routes the negative scores as in Section~\ref{sec:scan}, and the final prediction is $\hat{y} = \arg\max_c (\lambda\, s_c^+ + (1-\lambda)\,\hat{s}_c^-)$. All hyperparameters are selected by grid search on the validation set prior to test evaluation.

\section{Experimental Setup}
\vspace{-0.4em}

\begin{table*}[t]
\centering
\caption{Full numerical results on the few-shot learning task. For each dataset, we report the mean accuracy and 95\% confidence interval over 3 random seeds of our SCAN on 1-/2-/4-/8-/16-shot settings. We report the zero-shot performance of CLIP \cite{Radford2021LearningTV} for all settings. For TaskRes \cite{Yu2022TaskRF}, we report the results using the enhanced base classifier (i.e., TaskRes*). The best results are in \textbf{bold} and the second are \underline{underlined}.}
\vspace{0.6em}
\label{tab:main_results}
\resizebox{\textwidth}{!}{%
\begin{tabular}{l|c|cccccccccccc|c}
\toprule
\textbf{Method} & \textbf{Shot} & Caltech101\cite{FeiFei2004LearningGV} & DTD\cite{Cimpoi2013DescribingTI} & EuroSAT\cite{Helber2017EuroSATAN} & FGVCAir.\cite{Maji2013FineGrainedVC} & Flowers102\cite{Nilsback2008AutomatedFC} & Food101\cite{Bossard2014Food101M} & ImageNet\cite{Deng2009ImageNetAL} & Oxford Pets\cite{Parkhi2012CatsAD} & Stanford Cars\cite{Krause20133DOR} & SUN397\cite{Xiao2010SUNDL} & UCF101\cite{Soomro2012UCF101AD} & \textbf{Avg.} \\
\midrule

Zero-shot CLIP\cite{Radford2021LearningTV} & 0 & 84.52 & 40.33 & 41.80 & 16.98 & 65.46 & 77.31 & 60.33 & 85.51 & 54.26 & 58.56 & 61.44 & 58.77 \\

\midrule

CoOp\cite{Zhou2022ConditionalPL} & 1 & 87.43 & 44.13 & 50.51 & 9.80 & 67.90 & 73.71 & 57.15 & 86.51 & 55.48 & 60.10 & 62.10 & 59.53 \\
CoCoOp\cite{Zhou2022ConditionalPL} & 1 & 86.01 & 45.14 & 35.08 & 17.81 & 67.52 & 77.42 & 60.84 & \underline{86.96} & 57.22 & 62.28 & 62.84 & 59.92 \\
CLIP-Adapter\cite{Gao2021CLIPAdapterBV} & 1 & 88.70 & 46.66 & 61.51 & 17.21 & 73.43 & 76.77 & 61.20 & 85.99 & 55.14 & 61.28 & 62.29 & 62.65 \\
Tip-Adapter-F\cite{Zhang2022TipAdapterTA} & 1 & 89.38 & 50.31 & 59.16 & 20.83 & 80.13 & 77.61 & 61.32 & 86.47 & 58.51 & 62.51 & 64.91 & 64.65 \\
TaskRes\cite{Yu2022TaskRF} & 1 & 88.80 & 50.20 & 61.70 & 21.41 & 79.17 & 74.03 & 61.90 & 83.60 & 59.13 & 62.33 & 64.77 & 64.28 \\
SimNL\cite{Zhang2024EnhancingVF} & 1 & \underline{90.87} & \underline{53.13} & \underline{67.70} & \underline{23.61} & \underline{84.17} & \underline{77.93} & \underline{62.89} & 86.90 & \underline{61.34} & \underline{65.13} & \underline{68.25} & \underline{67.45} \\
SCAN (Ours) & 1 & \textbf{92.17} & \textbf{58.43} & \textbf{70.90} & \textbf{28.51} & \textbf{86.47} & \textbf{79.63} & \textbf{65.39} & \textbf{89.40} & \textbf{66.04} & \textbf{67.93} & \textbf{71.35} & \textbf{70.56} \\

\midrule

CoOp\cite{Zhou2022ConditionalPL} & 2 & 87.92 & 45.04 & 60.43 & 18.25 & 77.47 & 72.26 & 55.88 & 82.36 & 58.10 & 59.82 & 64.13 & 61.97 \\
CoCoOp\cite{Zhou2022ConditionalPL} & 2 & 89.47 & 46.20 & 38.51 & 20.22 & 70.70 & 78.81 & 61.86 & \underline{88.81} & 58.28 & 63.50 & 65.23 & 61.96 \\
CLIP-Adapter\cite{Gao2021CLIPAdapterBV} & 2 & 89.32 & 51.81 & 64.11 & 20.10 & 81.77 & 77.20 & 61.52 & 86.73 & 58.71 & 62.21 & 67.27 & 65.52 \\
Tip-Adapter-F\cite{Zhang2022TipAdapterTA} & 2 & 89.81 & 54.00 & 65.82 & 23.47 & 82.50 & 77.83 & 61.69 & 87.10 & 62.05 & 63.55 & 66.23 & 66.73 \\
TaskRes\cite{Yu2022TaskRF} & 2 & 90.27 & 55.13 & 65.83 & 24.13 & 86.57 & 75.17 & 62.63 & 84.63 & 63.70 & 64.97 & 70.00 & 67.54 \\
SimNL\cite{Zhang2024EnhancingVF} & 2 & \underline{91.19} & \underline{59.17} & \underline{74.40} & \underline{26.22} & \underline{88.43} & \underline{78.35} & \underline{63.47} & 87.68 & \underline{64.77} & \underline{66.73} & \underline{70.25} & \underline{70.06} \\
SCAN (Ours) & 2 & \textbf{92.69} & \textbf{65.47} & \textbf{78.00} & \textbf{31.92} & \textbf{91.03} & \textbf{80.85} & \textbf{66.27} & \textbf{90.68} & \textbf{69.87} & \textbf{69.83} & \textbf{74.35} & \textbf{73.72} \\

\midrule

CoOp\cite{Zhou2022ConditionalPL} & 4 & 89.17 & 53.38 & 70.20 & 21.72 & 85.81 & 72.72 & 59.91 & 87.22 & 61.92 & 63.46 & 67.08 & 66.60 \\
CoCoOp\cite{Zhou2022ConditionalPL} & 4 & 90.31 & 47.90 & 63.56 & 20.56 & 72.72 & \underline{79.51} & 62.52 & \underline{88.60} & 59.90 & 64.90 & 67.90 & 65.31 \\
CLIP-Adapter\cite{Gao2021CLIPAdapterBV} & 4 & 90.98 & 57.02 & 73.18 & 22.99 & 87.30 & 77.93 & 61.84 & 87.36 & 62.26 & 65.90 & 68.90 & 68.61 \\
Tip-Adapter-F\cite{Zhang2022TipAdapterTA} & 4 & 90.67 & 57.78 & 73.85 & 26.01 & 89.02 & 78.26 & 62.52 & 87.72 & 64.82 & 66.13 & 70.87 & 69.79 \\
TaskRes\cite{Yu2022TaskRF} & 4 & 90.97 & 60.70 & 73.83 & 25.70 & 90.20 & 76.10 & 63.57 & 86.33 & 67.43 & 67.27 & 70.93 & 70.28 \\
SimNL\cite{Zhang2024EnhancingVF} & 4 & \underline{92.21} & \underline{66.01} & \underline{76.54} & \underline{28.95} & \underline{92.04} & 78.74 & \underline{64.12} & 88.13 & \underline{67.96} & \underline{68.59} & \underline{73.46} & \underline{72.43} \\
SCAN (Ours) & 4 & \textbf{93.81} & \textbf{72.91} & \textbf{80.74} & \textbf{35.75} & \textbf{94.74} & \textbf{82.04} & \textbf{67.42} & \textbf{91.33} & \textbf{73.76} & \textbf{72.49} & \textbf{78.16} & \textbf{76.65} \\

\midrule

CoOp\cite{Zhou2022ConditionalPL} & 8 & 90.15 & 59.88 & 76.51 & 25.93 & 90.84 & 71.52 & 60.91 & 86.40 & 68.49 & 65.43 & 71.81 & 69.82 \\
CoCoOp\cite{Zhou2022ConditionalPL} & 8 & 90.14 & 52.21 & 64.13 & 22.03 & 75.88 & 79.59 & 62.40 & 88.74 & 60.87 & 65.37 & 68.25 & 66.33 \\
CLIP-Adapter\cite{Gao2021CLIPAdapterBV} & 8 & 91.22 & 60.70 & 78.34 & 25.77 & 91.79 & 78.01 & 62.68 & 87.70 & 67.78 & 67.52 & 72.02 & 71.32 \\
Tip-Adapter-F\cite{Zhang2022TipAdapterTA} & 8 & 91.54 & 62.67 & 77.83 & 30.21 & 91.85 & 78.71 & 64.00 & 88.07 & 69.53 & 68.80 & 74.50 & 72.52 \\
TaskRes\cite{Yu2022TaskRF} & 8 & 92.40 & 64.77 & 79.33 & 31.48 & 94.73 & 76.40 & 64.67 & 87.17 & 71.83 & 68.73 & 75.33 & 73.35 \\
SimNL\cite{Zhang2024EnhancingVF} & 8 & \underline{93.40} & \underline{67.78} & \underline{81.62} & \underline{33.90} & \underline{95.23} & \underline{79.23} & \underline{65.37} & \underline{89.29} & \underline{72.08} & \underline{70.93} & \underline{76.84} & \underline{75.06} \\
SCAN (Ours) & 8 & \textbf{95.00} & \textbf{75.28} & \textbf{86.62} & \textbf{41.40} & \textbf{97.03} & \textbf{82.93} & \textbf{68.87} & \textbf{92.29} & \textbf{79.28} & \textbf{75.43} & \textbf{82.34} & \textbf{79.68} \\

\midrule

CoOp\cite{Zhou2022ConditionalPL} & 16 & 91.61 & 63.11 & 82.36 & 31.01 & 84.39 & 73.80 & 62.95 & 87.30 & 72.51 & 69.11 & 75.70 & 73.07 \\
CoCoOp\cite{Zhou2022ConditionalPL} & 16 & 90.90 & 57.53 & 70.77 & 22.40 & 79.14 & 79.68 & 62.71 & 89.93 & 62.22 & 67.21 & 70.81 & 68.48 \\
CLIP-Adapter\cite{Gao2021CLIPAdapterBV} & 16 & 92.44 & 66.14 & 82.76 & 31.86 & 93.91 & 78.21 & 63.59 & 87.91 & 74.12 & 69.59 & 76.80 & 74.30 \\
Tip-Adapter-F\cite{Zhang2022TipAdapterTA} & 16 & 92.93 & 67.33 & 83.80 & 35.80 & 95.01 & 79.50 & 65.51 & 89.71 & 75.50 & 71.31 & 78.01 & 75.83 \\
TaskRes\cite{Yu2022TaskRF} & 16 & 93.43 & 67.13 & 84.03 & 36.30 & 96.03 & 77.60 & 65.73 & 87.83 & 76.83 & 70.67 & 77.97 & 75.78 \\
SimNL\cite{Zhang2024EnhancingVF} & 16 & \underline{93.77} & \underline{70.83} & \underline{87.36} & \underline{40.27} & \underline{96.51} & \underline{79.87} & \underline{66.52} & \underline{90.58} & \underline{77.48} & \underline{72.32} & \underline{80.28} & \underline{77.80} \\
SCAN (Ours) & 16 & \textbf{95.57} & \textbf{78.53} & \textbf{92.66} & \textbf{47.97} & \textbf{97.61} & \textbf{83.97} & \textbf{69.92} & \textbf{93.78} & \textbf{83.38} & \textbf{77.32} & \textbf{85.78} & \textbf{82.41} \\

\bottomrule
\end{tabular}
}
\end{table*}

We evaluate \textbf{SCAN} under two standard settings, few-shot learning and domain generalization.

\textbf{Few-Shot Learning.} We conduct experiments on 11 widely-used image classification benchmarks spanning diverse domains including general objects, fine-grained categories, textures, scenes, and remote sensing. These cover ImageNet~\cite{Deng2009ImageNetAL}, Caltech101~\cite{FeiFei2004LearningGV}, OxfordPets~\cite{Parkhi2012CatsAD}, StanfordCars~\cite{Krause20133DOR}, Flowers102~\cite{Nilsback2008AutomatedFC}, Food101~\cite{Bossard2014Food101M}, FGVC Aircraft~\cite{Maji2013FineGrainedVC}, DTD~\cite{Cimpoi2013DescribingTI}, SUN397~\cite{Xiao2010SUNDL}, EuroSAT~\cite{Helber2017EuroSATAN}, and UCF101~\cite{Soomro2012UCF101AD}.

\textbf{Domain Generalization.} To assess robustness under natural distribution shifts, we evaluate on four ImageNet variants, namely ImageNet-V2~\cite{Recht2019DoIC}, ImageNet-Sketch~\cite{Wang2019LearningRG}, ImageNet-A~\cite{Hendrycks2019NaturalAE}, and ImageNet-R~\cite{Hendrycks2020TheMF}. We also test resilience to label noise using a corrupted version of 16-shot ImageNet, where a fixed percentage of class labels are randomly flipped to simulate real-world annotation errors.


\textbf{Baseline Methods.} We compare SCAN against a range of vision-language adaptation methods including zero-shot and linear-probe CLIP~\cite{Radford2021LearningTV}, CoOp~\cite{Zhou2021LearningTP}, CoCoOp~\cite{Zhou2022ConditionalPL}, CLIP-Adapter~\cite{Gao2021CLIPAdapterBV}, Tip-Adapter-F~\cite{Zhang2022TipAdapterTA}, ProGrad~\cite{Zhu2022PromptalignedGF}, TPT~\cite{Shu2022TestTimePT}, TaskRes~\cite{Yu2022TaskRF}, and GraphAdapter~\cite{Li2023GraphAdapterTV}. We also compare directly with SimNL~\cite{Zhang2024EnhancingVF}, our primary baseline, which introduced dual positive-negative adapter routing. SCAN extends this foundation by introducing query-adaptive confusion routing at inference, LLM-bootstrapped contrastive prompts for finer textual discrimination, and a parameter-free adaptive fusion weight estimated from support-set geometry.

\section{Experimental Results and Analysis}

\begin{figure*}[t]
    \centering
    \includegraphics[width=\textwidth]{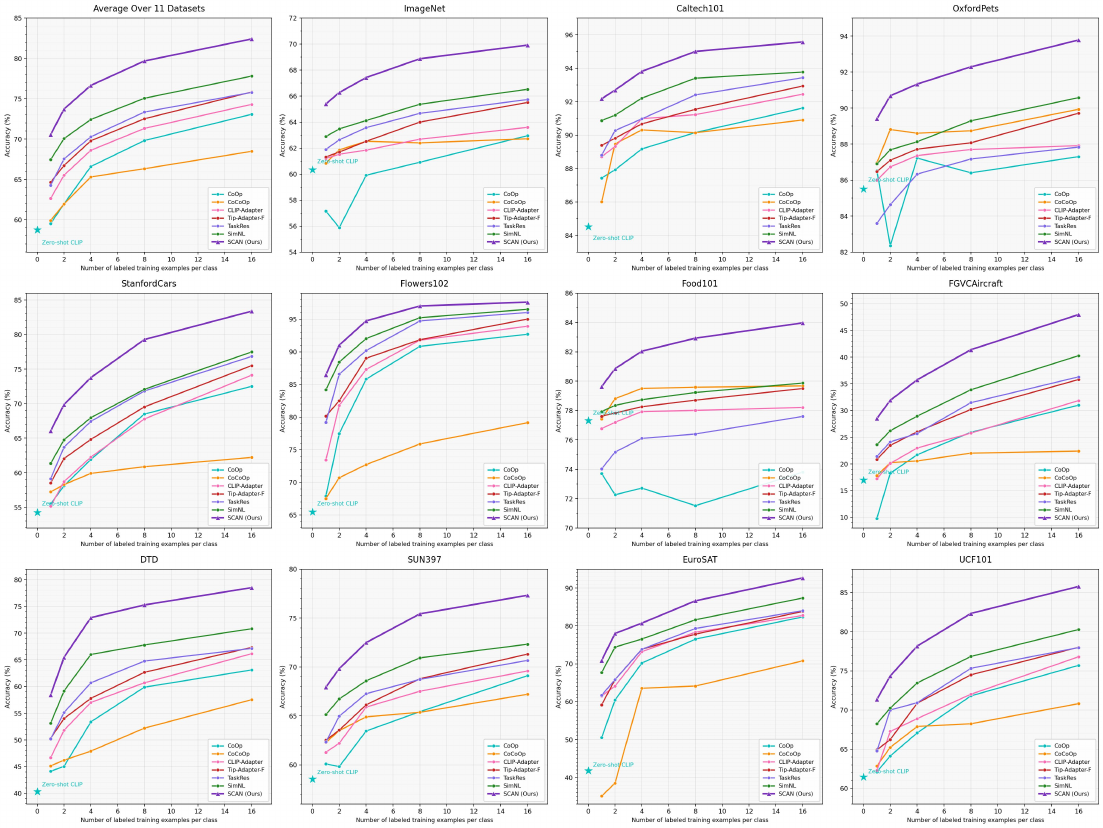}
    \vspace{-0.2cm}
    \caption{\textbf{Performance comparisons on few-shot learning on 11 image classification datasets.} For each dataset, we report the mean accuracy over 3 random seeds of our SCAN on 1-/2-/4-/8-/16-shot settings.}
    \label{fig:simnlpp_fewshot}
\end{figure*}

\vspace{-0.4em}
\noindent\textbf{Few-shot Learning.}
In Figure~\ref{fig:simnlpp_fewshot} and Table~\ref{tab:main_results}, SCAN consistently outperforms all baselines across 11 benchmarks and every shot setting. The largest gains appear on fine-grained and domain-specialized datasets, which is the expected signature of confusion-aware negative routing. DTD and FGVC Aircraft improve by 7.70\% over SimNL~\cite{Zhang2024EnhancingVF} at 16-shot, and Stanford Cars by 5.90\%, reflecting that suppressing confusable class signals has the greatest impact where inter-class visual similarity is highest. EuroSAT improves by 5.30\%, where style-conditioned prompts better distinguish spectrally similar land-cover categories. On ImageNet, SCAN achieves a 3.40\% improvement over SimNL at 16-shot, confirming that gains are non-marginal even on large-scale general-purpose classification. The average improvement over SimNL grows steadily from 3.11\% at 1-shot to 4.61\% at 16-shot, consistent with SCAN's routing mechanism benefiting from a richer support set as more labeled examples become available.

\begin{table*}[t]
    \centering
    \caption{\textbf{Robustness of SCAN to label noise and distribution shift.} Left: Accuracy on 16-shot ImageNet under different label corruption rates. SCAN consistently maintains stronger performance under noise than all baselines, including SimNL with plug-in reweighting. Right: Evaluation on target OOD variants of ImageNet (V2, Sketch, A, R). SCAN shows strong generalization under domain shift across all variants.}
    \vspace{0.8em}
    \begin{minipage}[t]{0.49\textwidth}
        \centering
        \scriptsize
        \setlength{\tabcolsep}{5pt}
        \renewcommand{\arraystretch}{1.1}
        \begin{tabular}{l|cccc}
            \toprule
            \textbf{Method} & \textbf{0\%} & \textbf{10\%} & \textbf{25\%} & \textbf{50\%} \\
            \midrule
            Tip-Adapter-F~\cite{Zhang2022TipAdapterTA} & 65.52 & 64.93 & 64.04 & 62.47 \\
            \quad + Reweighting & 65.64 & 65.25 & 64.55 & 63.39 \\
            \rowcolor{gray!10}
            \textit{Performance Gain} & +0.12 & +0.32 & +0.51 & +0.92 \\
            \midrule
            SimNL~\cite{Zhang2024EnhancingVF} & 66.31 & 65.54 & 64.77 & 63.37 \\
            \quad + Reweighting & 66.52 & 65.82 & 65.16 & 64.02 \\
            \rowcolor{gray!10}
            \textit{Performance Gain} & +0.21 & +0.28 & +0.39 & +0.65 \\
            \midrule
            \textbf{SCAN (Ours)} & \textbf{69.92} & \textbf{69.37} & \textbf{68.62} & \textbf{67.68} \\
            \bottomrule
        \end{tabular}
        \label{tab:label_noise}
    \end{minipage}
    \hfill
    \begin{minipage}[t]{0.50\textwidth}
        \centering
        \scriptsize
        \setlength{\tabcolsep}{1.5pt}
        \renewcommand{\arraystretch}{1.1}
        \begin{tabular}{l|cccccc}
            \toprule
            \textbf{Method} & \textbf{ImageNet} & \textbf{-V2} & \textbf{-Sketch} & \textbf{-A} & \textbf{-R} & \textbf{Avg.} \\
            \midrule
            Zero-Shot CLIP\cite{Radford2021LearningTV} & 60.33 & 53.27 & 35.44 & 21.65 & 56.00 & 41.59 \\
            LP CLIP\cite{Radford2021LearningTV} & 56.13 & 45.61 & 19.13 & 12.74 & 34.86 & 28.09 \\
            CoOp\cite{Zhou2022ConditionalPL} & 62.95 & 55.40 & 34.67 & 23.06 & 56.60 & 42.43 \\
            CoCoOp\cite{Zhou2022ConditionalPL} & 62.71 & 55.72 & 34.48 & 23.32 & 57.74 & 42.82 \\
            ProGrad\cite{Zhu2022PromptalignedGF} & 62.17 & 54.70 & 34.40 & 23.05 & 56.77 & 42.23 \\
            TPT\cite{Shu2022TestTimePT} & 60.74 & 54.70 & 35.09 & \underline{26.67} & 59.11 & 43.89 \\
            TaskRes\cite{Yu2022TaskRF} & 64.75 & 56.47 & 35.83 & 22.80 & 60.70 & 43.95 \\
            GraphAdapter\cite{Li2023GraphAdapterTV} & 64.94 & 56.58 & 35.89 & 23.07 & 60.86 & 44.10 \\
            SimNL\cite{Zhang2024EnhancingVF} & \underline{66.52} & \underline{57.87} & \underline{36.38} & 25.73 & \underline{61.12} & \underline{45.28} \\
            \rowcolor{gray!10}
            \textbf{SCAN (Ours)} & \textbf{69.92} & \textbf{60.57} & \textbf{39.58} & \textbf{28.53} & \textbf{64.22} & \textbf{48.23} \\
            \bottomrule
        \end{tabular}
        \label{tab:domain_generalization}
    \end{minipage}
\end{table*}

\noindent\textbf{Robustness to Natural Distribution Shifts.}
Table~\ref{tab:domain_generalization} (\textit{Right}) shows SCAN achieving the highest accuracy across all four ImageNet OOD variants, with an average gain of 2.95\% over SimNL~\cite{Zhang2024EnhancingVF}. The strongest improvement appears on ImageNet-Sketch (+3.20\%), where texture and style cues diverge most from the source domain, suggesting that SCAN's style-conditioned prompts and LLM-generated contrastive descriptions help retain semantic alignment when visual appearance shifts substantially. Gains on ImageNet-A (+2.80\%) and ImageNet-R (+3.10\%) further confirm that confusion-aware routing remains effective under adversarial and artistic domain shifts. These results indicate that SCAN's improvements are not confined to the source distribution but transfer consistently across diverse shift types.

\noindent\textbf{Robustness to Label Noise.}
Table~\ref{tab:label_noise} (\textit{Left}) reports accuracy on 16-shot ImageNet~\cite{Deng2009ImageNetAL} under 0--50\% label corruption. SCAN not only starts from a higher clean baseline (69.92\%) but degrades more gracefully than SimNL~\cite{Zhang2024EnhancingVF} under increasing noise, dropping only 2.24 points at 50\% corruption compared to 2.50 for SimNL with plug-in reweighting. Notably, SCAN under 50\% label noise (67.68\%) still exceeds SimNL's clean performance (66.52\%), reinforcing that query-adaptive routing is intrinsically noise-tolerant. This robustness arises because SCAN routes negatives based on text-side confusion signals rather than visual cache similarity alone, so corrupted support labels have a reduced effect on inference-time decisions.

\vspace{-0.8em}
\section{Conclusion and Future Work}
\vspace{-1em}
We introduced SCAN, a few-shot vision-language adaptation framework that addresses three core limitations of prior negative-space methods through query-adaptive confusion routing, LLM-bootstrapped contrastive prompts, and a parameter-free adaptive fusion weight estimated from support-set geometry. SCAN consistently outperforms strong baselines across 11 benchmarks and four OOD variants, with the largest gains on fine-grained datasets where confusable class suppression is most consequential, and remains robust under significant label noise. These results demonstrate that targeted improvements to the negative branch and textual supervision can yield substantial and consistent gains without increasing inference-time parameters. In future work, we plan to extend query-adaptive routing to open-vocabulary detection and segmentation, explore its applicability in continual few-shot settings, and investigate scaling to larger vision-language backbones.

\bibliography{egbib}
\end{document}